\documentclass[letterpaper, 10 pt, conference]{support/ieeeconf}
\usepackage{times}
\usepackage[pdftex]{graphicx}
\usepackage{amsmath,amssymb,amsopn,amstext,amsfonts}
\usepackage{cancel}
\usepackage[space]{cite}
\usepackage{pdfsync}
\usepackage{balance}
\usepackage{color}
\usepackage{mathtools}
\usepackage[ruled,vlined]{algorithm2e}
\usepackage{bm}

\usepackage{diagbox}
\usepackage{float}
\usepackage{epstopdf}
\usepackage{pifont}
\usepackage{fixltx2e}
\usepackage{amsmath}
\usepackage{multirow}
\usepackage{booktabs}
\usepackage{url}
\usepackage{svg}
\usepackage{threeparttable}
\usepackage[linkcolor=black,citecolor=black,urlcolor=black,colorlinks=true]{hyperref}
\usepackage{subcaption}
\usepackage{slashbox}
\usepackage{makecell}

\newcommand{\tp}{^{\mathrm{T}}}

\graphicspath{{./figures/}}
\DeclareGraphicsExtensions{.png,.jpg,.eps,.pdf}
\IEEEoverridecommandlockouts
\makeatletter
\def\endthebibliography{%
	\def\@noitemerr{\@latex@warning{Empty `thebibliography' environment}}%
	\endlist
}

\title{\LARGE \bf
Robo-centric ESDF: \\A Fast and Accurate Whole-body Collision Evaluation Tool \\for Any-shape Robotic Planning
}

\author{Shuang Geng $^{\dag \,}$\textsuperscript{1,2}, 
   Qianhao Wang $^{\dag \,}$\textsuperscript{1,2}, 
	Lei Xie \textsuperscript{1}, 
	Chao Xu \textsuperscript{1,2}, 
	Yanjun Cao \textsuperscript{1,2},  
	and Fei Gao\textsuperscript{1,2}
	\thanks{\textbf{${\dag}$ Equal contribution.}}        
	\thanks{1 Zhejiang University, Hangzhou, 310013, China.} 
	\thanks{2 Huzhou Institute of Zhejiang University, Huzhou, 313000, China.}
	\thanks{Email:{\tt\footnotesize \{gengshuang, qhwangaa, fgaoaa\}@zju.edu.cn}}
	\thanks{Corresponding Author: Fei Gao and Qianhao Wang}
}

\begin{document}

\makeatletter
 \let\@oldmaketitle\@maketitle
 \renewcommand{\@maketitle}{\@oldmaketitle
  \includegraphics[width=\linewidth]{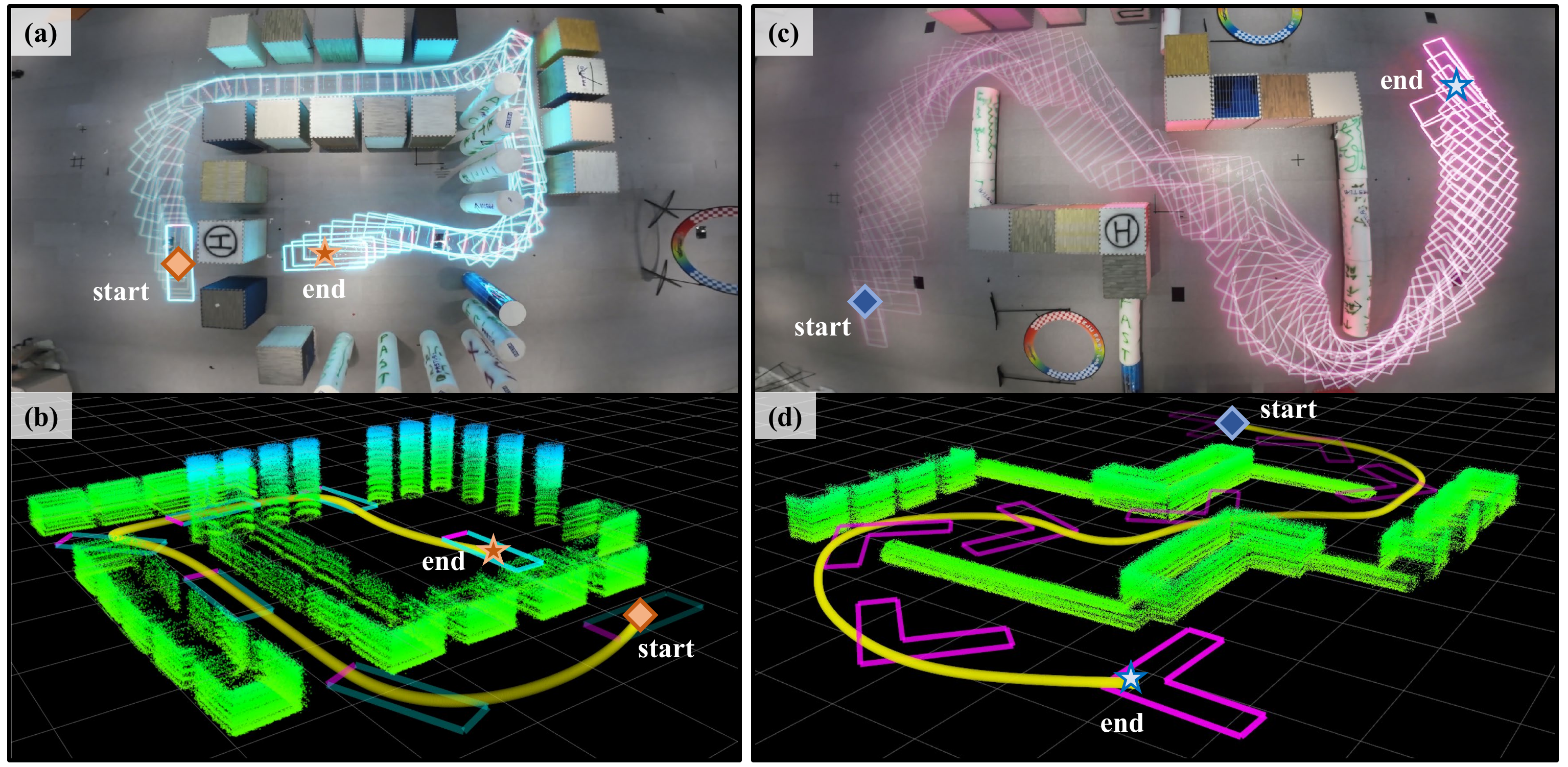}
  \captionsetup{font={small}}
  \captionof{figure}{ \label{fig:top}
  Real-world experiments for rectangle-shape robot and L-shape robot to pass through narrow channels and gaps. (a): BEV(Bird's Eye View) of the rectangle-shape robot real-world experiment, pink line indicates the front of the robot. (b): The trajectory of the rectangle-shape robot is visualized in a point cloud map. (c): BEV of L-shape robot real-world experiment. (d):The trajectory of the L-shape robot is visualized in a point cloud map.
  }
  \vspace{0.4cm} 
 }
 \makeatother
 \maketitle
  \setcounter{figure}{1}
  

\thispagestyle{empty}
\pagestyle{empty}


\begin{abstract} 
   For letting mobile robots travel flexibly through complicated environments, increasing attention has been paid to the whole-body collision evaluation. 
   Most existing works either opt for the conservative corridor-based methods that impose strict requirements on the corridor generation, or ESDF-based methods that suffer from high computational overhead. It is still a great challenge to achieve fast and accurate whole-body collision evaluation.
   In this paper, we propose a Robo-centric ESDF (RC-ESDF) that is pre-built in the robot body frame and is capable of seamlessly applied to any-shape mobile robots, even for those with non-convex shapes. 
   RC-ESDF enjoys lazy collision evaluation, which retains only the minimum information sufficient for whole-body safety constraint and significantly speeds up trajectory optimization.
   Based on the analytical gradients provided by RC-ESDF, we optimize the  position and rotation of robot jointly, with whole-body safety, smoothness, and dynamical feasibility taken into account.
   Extensive simulation and real-world experiments verified the reliability and generalizability of our method. 
\end{abstract}


\section{Introduction}
\label{sec:Introduction}
Collision evaluation is a core module of mobile robot trajectory optimization, ensuring that robot can move safely in narrow environment. Commonly, a simplified model for robot collision evaluation, for instance, modeling a robot as a mass point with linear dynamics \cite{zhou2020egoplanner, zhou2019robust, gao2020teach}, is sufficient for simple tasks. 
However, for precisely moving among dense obstacles, the shape of a robot, especially for non-convex shape, needs to be explicitly considered in collision evaluation. 
Generating a trajectory with a robot's shape, namely whole-body planning, not only requires careful consideration of robot's position, but also its rotation.
This demands much more complicated computation, making whole-body collision evaluation hard to meet limited onboard resources in practice.
How to efficiently generate collision-free trajectories for an any-shape  robot in real-time is still a great challenge.

For whole-body collision evaluation, apart from some task-specific methods \cite{ji2021mapless, liu2018search}, the most popular options are to use GJK-based\cite{zhang2018autonomous,he2021tdr}, corridor-based \cite{han2021fast,ding2019safe,manzinger2020using} or ESDF-based \cite{zhou2019robust, li2021optimization} methods.
GJK-based methods require complex pre-processing to model obstacles as convex polyhedrons. 
Corridor-based methods constrain the mobile robot inside the corridor that composed of several consecutive convex regions. However, the corridor generation tightly requires that the intersection of its adjacent components have to contain at least one robot. When applied to narrow spaces and non-convex robots, they turn to have no feasible solution, as shown in Fig. \ref{fig:l-compare}(c).
ESDF-based methods need to maintain a cost field map, yet suffer from a dilemma between the computational overhead and the field range.
Either some ESDF-based methods model the robot conservatively as a combination of circles to achieve fast convergence but optimize failed in narrow environment, as shown in Fig. \ref{fig:benchmark}(b), or others densely sample on the robot to ensure accuracy of collision evaluation but take long calculation time, as illustrated in Fig. \ref{fig:esdf_build_time} and Table \ref{tab:table 1}.

Considering the limitations of current methods, we summarize the following main challenges for an ideal whole-body collision evaluation: 
\textbf{(1)}~Efficiency: low computational overhead is preferred.
\textbf{(2)}~Accurate shape modeling: when applied in narrow scenarios, the method should avoid the failure due to the low-fidelity  approximation introduced by conservative modeling, as shown in Fig. \ref{fig:benchmark}(b).
\textbf{(3)}~Generality: the method is supposed to be readily applicable to  robots of any-shape, even non-convex robots, such as ground robots and aerial robots mentioned in Sec.\ref{sec:Experiment Results} .

\begin{figure}[!t]
	\centering
   \vspace{0.5cm}
	\includegraphics[width=1\linewidth]{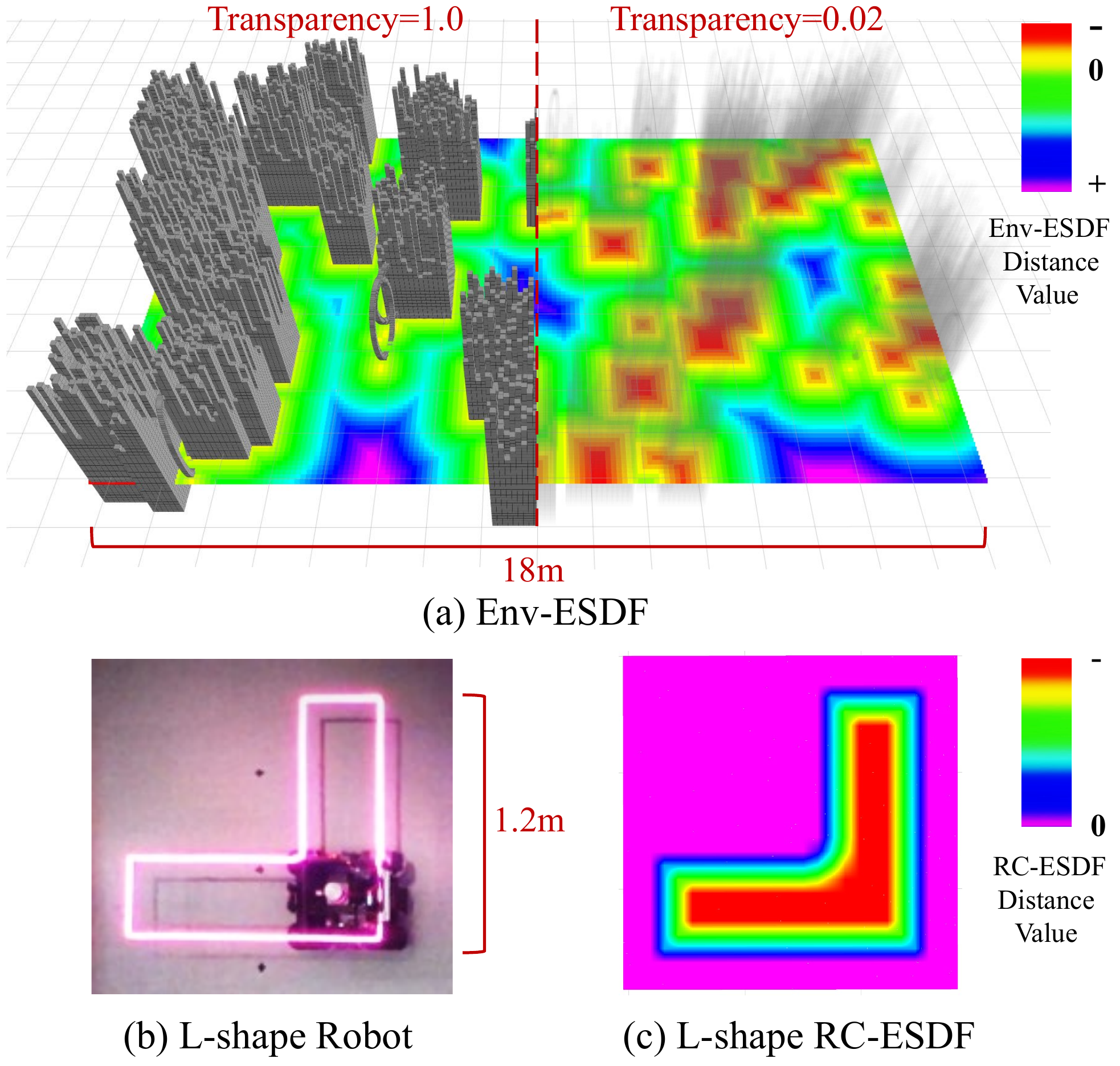}
   \captionsetup{font={small}}
	\caption{
      Comparison of RC-ESDF and Env-ESDF. (a): Env-ESDF built on $18m\times10m$ point cloud map. The transparency of obstacles on the left side is 1.0, which clearly shows the shape of obstacles. The transparency of the obstacle on the right side of the image is 0.02. Colors indicate the change of ESDF value of the Env-ESDF.  ESDF values inside the obstacles are negative, and outside are positive. (b): L-shape robot with $1.2m$ in length for real-world experiments. (c): RC-ESDF is built based on L-shape robot. ESDF values inside the robot are negative, and outside are zero.
	}
    \vspace{-0.6cm}
	\label{fig:filed_compare}
\end{figure}

To bridge this gap, we propose a Robo-centric Euclidean Signed Distance Field  (RC-ESDF) to achieve fast and accurate whole-body collision evaluation.
RC-ESDF is built offline in the robot body frame, whose shape and size are similar as the robot, as shown in Fig. \ref{fig:filed_compare}(c), and it does not require a real-time update. Moreover, RC-ESDF enjoys lazy collision evaluation that only focus on the obstacle points collide with robot, effectively reducing the computational overhead as shown in Table \ref{tab:table 1}. We jointly optimize the robot's position and rotation, both convex and non-convex shape mobile robots obtain better trajectory optimization results in narrow environments, as presented in Fig. \ref{fig:benchmark} and \ref{fig:l-compare}.
To verify the generalizability and reliability of our method, we implement extensive experiments in simulation and real-world environments with rectangle-shape (convex, as shown in Fig. 1(a)) and L-shape (non-convex, as shown in Fig. 1(c)) mobile robots.
The main contributions of this paper are:
\begin{itemize}
   \item We propose a novel RC-ESDF for whole-body collision evaluation, reducing considerable computational overhead as it is pre-built without the need for real-time updating and enjoys lazy collision check.
   \item Based on the RC-ESDF, we jointly optimize the position and rotation of robot simultaneously in narrow environments.
   \item We conduct extensive simulation and real-world experiments to validate that our method accurately describes both convex and non-convex shape robots and generates a collision-free trajectory.
\end{itemize}


\section{Related Works}
\label{sec:relatedworks}
There are various planning works \cite{zhou2020egoplanner, zhou2019robust} that model a robot as a mass point and inflate obstacles based on the robot's radius, which achieve collision evaluation by ensuring the center of the sphere in the free space of the inflated map.
However, this conservative method, which does not take the specific shape of the robot into account, is not applicable in crowded environments.
Ji et al. \cite{ji2022real} propose to model the drone as a disc for perching on a moving platform. 
To cross narrow gaps, Liu et al. \cite{liu2018search} use a search-based method that formulates the robot as an ellipsoid and checks if the ellipsoid collides with obstacles along every primitive. 
In summary, these simple modeling methods are not accurate enough to generally represent any-shape mobile robots.


The most popular whole-body collision evaluation methods for mobile robot are GJK-based, corridor-based and ESDF-based method.
GJK-based methods\cite{gilbert1988fast, cameron1997enhancing,zhang2018autonomous,he2021tdr} that represent the robot and obstacles using convex polyhedron and constructs safety constraint by calculating the distance between two convex polyhedrons. However, this method need to ensure that all obstacles are convex-shape, and the solution dimension of the problem increases significantly as the number of obstacles increases.

There are a number of frameworks that opt for corridor-based methods to achieve whole-body collision evaluation.
Generally, corridor is constructed by a set of polyhedron-shape \cite{wang2022geometrically,han2021fast,gao2020teach}, sphere-shape \cite{ji2021mapless,zhu2015convex} , or rectangle-shape\cite{ding2019safe, manzinger2020using} convex hulls.
Han et al. \cite{han2021fast} propose a baseline for autonomous drone racing, which takes the drone's shape into account to fly through narrow gaps. This method models the robot as a convex polyhedron, then constrains the convex polyhedron in the flight corridor to guarantee safety.
Ding \cite{ding2019safe} and Manzinger \cite{manzinger2020using} construct a series of rectangles as a safe corridor for car-like robot. Although this method speed up the construction of the corridor, it sacrifices a significant amount of solution space.
However, the above corridor-based methods strictly require that the intersection of adjacent compositions of the corridor can contain at least one robot. Worse, these methods are naturally too conservative for non-convex shape robots, such as the L-shape robot, as shown in Fig. \ref{fig:benchmark}(c) and \ref{fig:l-compare}(c).

ESDF-based method is also heavily used in collision evaluation\cite{zhou2019robust} as a map representation that easily provides the distance to the nearest obstacle, which we call environment-based ESDF (abbreviated as Env-ESDF) in this paper.
Some methods model the robot with a series of circles\cite{li2021optimization,li2021optimal}.
Based on Env-ESDF, safety can be achieved by constraining the distance between the center of the circles and the obstacles to be greater than the circles' radius.
However, this conservative method is struggling to deal with the narrow environment, as shown in Fig. \ref{fig:benchmark}(b).
An intuitive idea for ESDF-based whole-body collision evaluation method is dense sampling on the robot, which can accurately describe the shape of robot, but requires more complex calculations, as shown in Fig. \ref{fig:esdf_build_time} and Table \ref{tab:table 1}.
Few methods can satisfy both accuracy and low computational consumption.  
Moreover, we need to build a large Env-ESDF to ensure that the trajectory is always in the field, thus the trade-off between the range of the ESDF and the computational overhead of building the field keeps a challenging problem.


\section{Robo-centric ESDF}
\label{sec:ROBO-CENTRIC ESDF}
In this section, we present the details of constructing RC-ESDF. It is pre-built in the robot body frame, which only focuses on the shape of the robot rather than the obstacles in the environment. 
Compared to Env-ESDF, RC-ESDF does not require real-time updates based on the environment information obtained from sensors, which greatly reduces the computational overhead. 
In the same way that Env-ESDF can describe obstacles of any-shape, RC-ESDF has natural applicability to represent any-shape robot, even for non-convex shape robots.
As shown in Fig. \ref{fig:filed_compare}(c), RC-ESDF accurately models the robot according to its shape. 

We define ESDF values inside the robot are negative, and its norm is the closest distance to the robot surface.
ESDF values outside the robot are zero, which ensures that there are no discontinuities in the optimization due to the field with abrupt boundaries.
In contrast, Env-ESDF based methods can only ensure continuity for optimization by creating a larger field.
This method ignores the obstacles information out of RC-ESDF that does not collide with the robot, and only evaluates collision by the ESDF values of the obstacle points that intersect with RC-ESDF. 
For the calculation of ESDF values, RC-ESDF utilizes an efficient $O(n)$ algorithm \cite{felzenszwalb2012distance}. 
This algorithm requires RC-ESDF to be stored by a grid map. 
And we store ESDF value of the vertices of the grids.
ESDF values and gradient information of any obstacle point in RC-ESDF can be obtained by linear interpolation \cite{zhou2019robust} of the ESDF values of the saved vertices.

\begin{figure}[!t]
    \centering 
    \vspace{0.1cm}
    \includegraphics[width=1\linewidth]{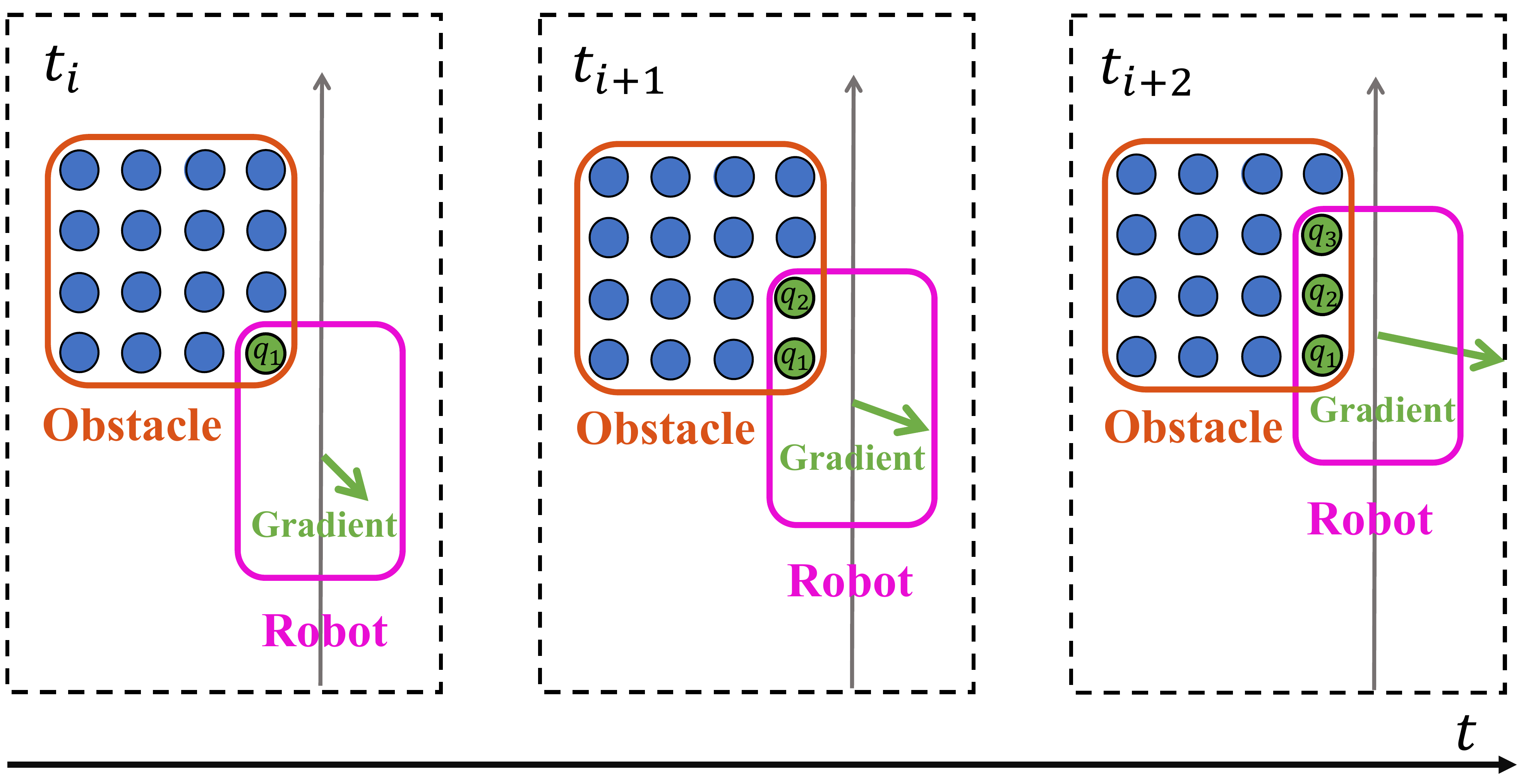}
    \captionsetup{font={small}}
    \caption{
    This figure illustrates the process of the robot colliding with an obstacle in the world frame. 
    The gray arrow indicates the direction of the robot's movement. 
    The green point $q_i$ is the obstacle point colliding with the robot. 
    The green arrow represents the norm and direction of the sum of the gradient that obstacle points generated in RC-ESDF.  }
    \label{fig:world-frame}
\end{figure}

\begin{figure}[!t]
    \centering 
    \vspace{-0.3cm}
    \includegraphics[width=1\linewidth]{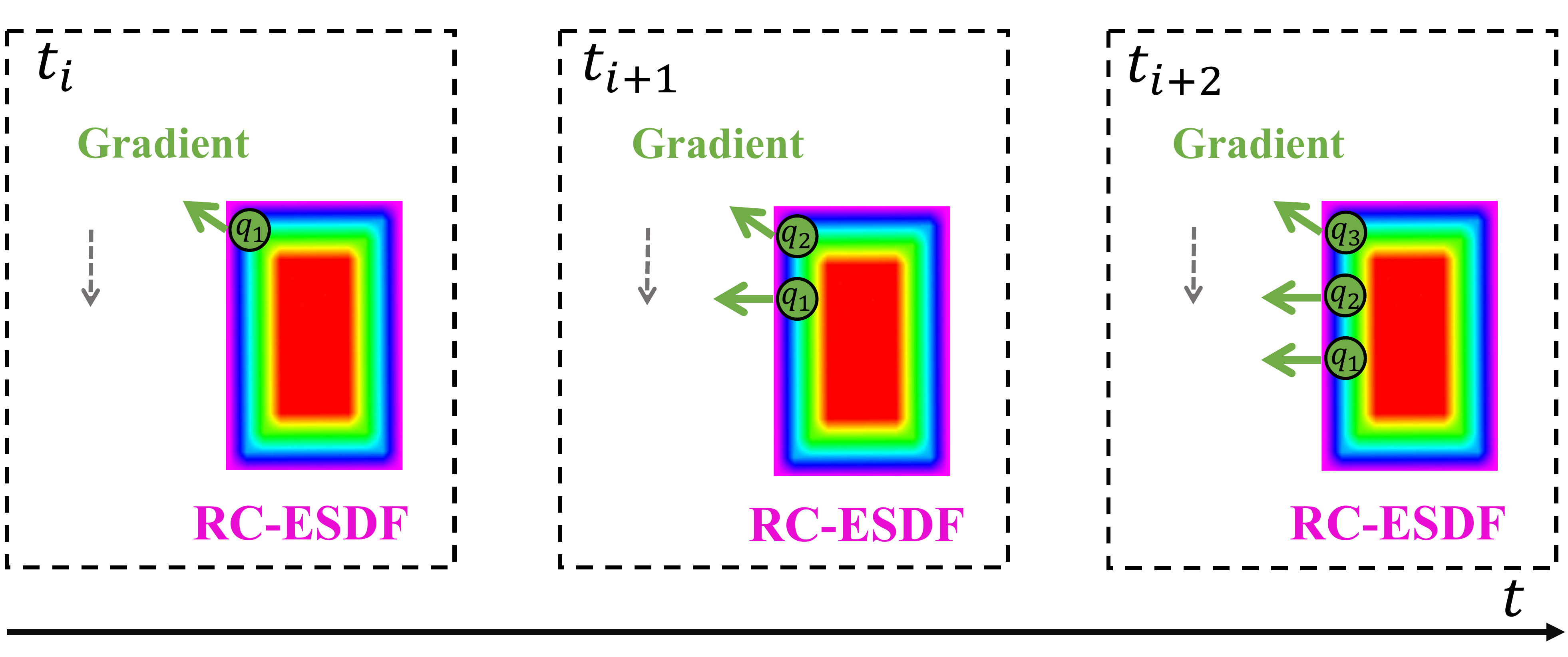}
    \captionsetup{font={small}}
    \caption{
    This figure illustrates the process of the robot colliding with an obstacle in the body frame. 
    Movement of the green obstacle points intersect with the robot along the gray dashed arrows in RC-ESDF.  The green arrows represent the gradient of obstacle points generated in RC-ESDF.}
    \vspace{-1.2cm}
    \label{fig:body-frame}
\end{figure}

To clearly illustrate the gradient of collision evaluation based on RC-ESDF, we show the movement of the robot intersecting with an obstacle as time changes in Fig. \ref{fig:world-frame} and  \ref{fig:body-frame}.
In this demonstration, the obstacle is represented by several blue points and collision obstacle points have green color.
In the world frame, as illustrated in Fig. \ref{fig:world-frame}, the robot gradually moves into the range of the obstacle following the direction of the solid gray arrow. 
Meanwhile, in the robot body frame, as shown in  Fig. \ref{fig:body-frame}, the obstacle points progressively move into the RC-ESDF. 
Thus we can obtain ESDF values and gradients of all the obstacle points in RC-ESDF at different time $t$. 
Conducting gradients to the control points of trajectory through the chain rule, the joint safety constraint on position and rotation is constructed. The detailed calculation process is in  Sec.\ref{sec:collision_penalty}. 

\section{Gradient-based Joint Trajectory Optimization}

\subsection{Trajectory Representation}

We jointly optimize the position and yaw of robot simultaneously. 
It can generate a whole-body collision-free trajectory considering smoothness and feasibility. 
For a holonomic ground-robot, its trajectory is represented by $\left\{x(t), y(t), \psi(t)\right\} \in SE(2)$. 
The trajectory is parameterized by a uniform B-spline curve, which is a piecewise polynomial uniquely determined by its degree $p_b$, a knot span $\Delta t$, and $N_c$ control points $\left\{\mathbf{Q}_k, {\Psi}_k\right\}$. 
In practice, we choose $p_b = 3$, then the total duration of the trajectory is $(N_c-3)\Delta t$. 
The control points of the velocity, acceleration, and jerk curves can be obtained by:

\begin{equation}
\begin{aligned}
\begin{split}
\mathbf{V}_k=\frac{\mathbf{Q}_{i+k}-\mathbf{Q}_k}{\Delta t},&\mathbf{A}_k=\frac{\mathbf{V}_{k+1}-\mathbf{V}_k}{\Delta t},\mathbf{J}_k=\frac{\mathbf{A}_{k+1}-\mathbf{A}_k}{\Delta t},\\
&k\in \left\{1, 2, ... , N_c-2\right\}.
\end{split}
\end{aligned}    
\end{equation}

 The control points of yaw velocity ${V}_{\Psi,k}$, yaw acceleration ${A}_{\Psi,k}$, and yaw jerk ${J}_{\Psi,k}$ have the same expression. We optimize the subset of $N_c+1-2p_b$ control points $\left\{\mathbf{Q}_{p_b}, \mathbf{Q}_{p_b+1}, . . . , \mathbf{Q}_{N_c - {p_b}} \right\} $ and $\left\{{\Psi}_{p_b}, {\Psi}_{p_b+1}, . . . , {\Psi}_{N_c - {p_b}} \right\}$. The first and last $p_b$ control points should not be changed because they determine the boundary state.

\subsection{Objective Functions}

The optimization problem is formulated as follows:
\begin{equation}
\min_{\mathbf{Q},\Psi} J = \lambda_{ps}J_{ps}+\lambda_{pf}J_{pf}+\lambda_{\psi s}J_{\psi s}+\lambda_{\psi f}J_{\psi f}+\lambda_{c}J_{c},
\end{equation}
where $J_{ps}$ and $J_{pf}$ are the smoothness and feasibility penalty of position, $J_{\psi s}$ and $J_{\psi f}$ are the smoothness and feasibility penalty of yaw angle, $J_c$ is the joint collision penalty associated with both position and yaw. 
$\lambda_{ps}$, $\lambda_{pf}$, $\lambda_{\psi s}$, $\lambda_{\psi f}$, $\lambda_{c}$ are weights for each penalty terms.

\subsubsection{Collision penalty}
\label{sec:collision_penalty}
Collision penalty pushes the entire robot away from obstacles at each constraint point $[\mathbf{p}_k^\mathrm{T}, \psi_k]$ to ensure that the whole trajectory is collision-free. According to the properties of the third-order B-spline, constraint point $\mathbf{p}_k$ and $\psi_k$ is defined as: 

\begin{equation}
\begin{aligned}
\begin{split}
&\mathbf{p}_k = \frac{1}{6}({\mathbf{Q}_k+4\mathbf{Q}_{k+1}+\mathbf{Q}_{k+2}}),\\
&\psi_k = \frac{1}{6}({{\Psi}_k+4{\Psi}_{k+1}+{\Psi}_{k+2}}),\\
\end{split}
\end{aligned}    
\end{equation}
where $\mathbf{p}_k = [x_{k}, y_{k}]^\mathrm{T}$ is the center of rotation of the robot in the world frame, and $\psi_k$ is yaw angle. We define collision penalty $J_c$ as:

\begin{equation}
\begin{aligned}
\begin{split}
&J_c =  \sum_{k=1}^{N_c-2} F_c(\mathbf{p}_k,\psi_k),\\
&F_c(\mathbf{p}_k,\psi_k) = d_{\mathbf{p}_k,\psi_k}^2,
\end{split}
\end{aligned}
\end{equation}
where $F_c(\mathbf{p}_k,\psi_k)$ is  a differentiable potential cost function. $d_{\mathbf{p}_k,\psi_k}$ is the sum of RC-ESDF values of obstacle points that collide with robot whose pose is $[\mathbf{p}_k^\mathrm{T}, \psi_k]$:

\begin{equation}
\begin{aligned}
\begin{split}
d_{\mathbf{p}_k,\psi_k} = \sum_{i=1}^{M_d} d_i,
\forall i &\in \left\{1, 2,  ..., M_d\right\},
\end{split}
\end{aligned}
\end{equation}
where $M_d$ is number of collision points, $d_i$ is the RC-ESDF value of $i$-th collision point. $d_i$ can easily be obtained from RC-ESDF, which is mentioned in Sec.\uppercase\expandafter{\romannumeral3}.

We define the $j$-th vertex $\mathbf{v}_{k,j}$ of the RC-ESDF in the world frame as:
\begin{equation}
\mathbf{v}_{k,j} = \mathbf{R}_k \mathbf{v}_{b,j}+\mathbf{p}_k \ \forall j \in \left\{1,2,...,M_v\right\},
\end{equation}
where $M_v$ is the number of vertexes and $\mathbf{v}_{b,j}=[x_{b,j}, y_{b,j}]^\mathrm{T}$ is the $j$-th vertex of RC-ESDF in the robot body frame. Note $\mathbf{v}_{b,j}$ is constant once the robot's shape is identified. $\mathbf{R}_k$ is rotation matrix that represents the robot's rotation. 
We use Axis-aligned bounding box (AABB) algorithm to delineate a rectangular area based on where the robot is located in the world frame. 
The point $\mathbf{q}_{w,i}$ of obstacles in the AABB bounding box, which are defined as collison point, can be transferred to $\mathbf{q}_{b,i}$ in the body frame :

\begin{equation}
\begin{aligned}
\begin{split}
\mathbf{q}_{b,i} = \mathbf{R}_k^{-1}(\mathbf{q}_{w,i}-\mathbf{p}_k).
\end{split}
\end{aligned}
\end{equation}

We use $H_{i,k}$ to represent the RC-ESDF value of the $i$-th collision point $\mathbf{q}_{b,i}$ when the robot is at $[\mathbf{p}_k^\mathrm{T}, \psi_k]$.
The gradients of the RC-ESDF value $H_{i,k}$ with respect to $\mathbf{p}_k$ and $\psi_k$ can be calculated by:
\begin{equation}
\begin{aligned}
\begin{split}
&\frac{\partial H_{i,k}}{\partial \mathbf{p}_k} = 
\left( \frac{\partial \mathbf{q}_{b,i}}{\partial \mathbf{p}_k} \right)\tp 
\frac{\partial H_{i,k}}{\partial \mathbf{q}_{b,i}}=
-\mathbf{R}_k \frac{\partial H_{i,k}}{\partial \mathbf{q}_{b,i}},\\
&\frac{\partial H_{i,k}}{\partial \psi_k} = 
\left( \frac{\partial \mathbf{q}_{b,i}}{\partial \psi_k} \right)\tp
\frac{\partial H_{i,k}}{\partial \mathbf{q}_{b,i}}\\
&=
(\mathbf{q}_{w,i}-\mathbf{p}_k)\tp
\left[ \begin{array}{cc}-sin\psi_k & -cos\psi_k \\ cos\psi_k & -sin\psi_k \end{array} \right]
{\frac{\partial H_{i,k}}{\partial \mathbf{q}_{b,i}}}.\\
\end{split}
\end{aligned}
\end{equation}

Then the gradient of the cost function  $F_c(\mathbf{p}_k,\psi_k)$ with respect to $\mathbf{p}_k$ and $\psi_k$ can be written as:

\begin{equation}
\begin{aligned}
\begin{split}
\frac {\partial F_c(\mathbf{p}_k, \psi_k)}{\partial \mathbf{p}_k} &= 2d_{\mathbf{p}_k,\psi_k}\sum_{i=1}^{M_d} \frac{\partial H_{i,k}}{\partial \mathbf{p}_k},\\
\frac {\partial F_c(\mathbf{p}_k, \psi_k)}{\partial \psi_k} &= 2d_{\mathbf{p}_k,\psi_k}\sum_{i=1}^{M_d} \frac{\partial H_{i,k}}{\partial \psi_k}.
\end{split}
\end{aligned}
\end{equation}

\subsubsection{Feasibility penalty}
Feasibility penalty constrains the velocity and acceleration along the trajectory from the exceeding maximum value $v_{m}$ and $a_{m}$, we define the feasibility penalty function $J_{pf}$ as:

\begin{equation}
J_{pf} = \sum_{k=1}^{N_c-1} (||\mathbf{V}_k||^2-v_{m}^2)+\sum_{k=1}^{N_c-2} (||\mathbf{A}_k||^2-a_{m}^2).
\vspace{1.0cm}
\end{equation}

To limit the velocity and acceleration of yaw angle, the yaw feasibility penalty function $J_{\psi f}$ can be written as:

\begin{equation}
J_{\psi f} = \sum_{k=1}^{N_c-1} (||{V}_{\psi,k}||^2-{v}_{\psi,m}^2)+\sum_{k=1}^{N_c-2} (||{A}_{\psi,k}||^2-a_{\psi,m}^2).
\end{equation}

{\setlength{\parindent}{0cm} where $\mathbf{v}_{\psi,m}$ and $a_{\psi,m}$ are the maximum value of yaw velocity and acceleration. }

\subsubsection{Smoothness penalty}
In the paper, minimizing the control points of second and third-order derivatives of the B-spline trajectory is sufficient to reduce the derivative along the whole curve. We use squared acceleration and jerk as smoothness penalty function $J_{ps}$ and $J_{\psi s}$:
\begin{equation}
\begin{aligned}
\begin{split}
J_{ps}&=\sum_{k=1}^{N_c-1} ||\mathbf{A}_k||_2^2 + \sum_{k=1}^{N_c-2} ||\mathbf{J}_k||_2^2,\\
J_{\psi s}&=\sum_{k=1}^{N_c-1} ||{A}_{\psi,k}||_2^2 + \sum_{k=1}^{N_c-2} ||{J}_{\psi,k}||_2^2.
\end{split}
\end{aligned}
\end{equation}

\subsection{Numerical Optimization}

We adopt L-BFGS\footnote{https://github.com/ZJU-FAST-Lab/LBFGS-Lite} \cite{liu1989limited} to solve the numerical problem in trajectory optimization.
Since the ESDF values are obtained by linear interpolation, the gradient of the ESDF values with respect to the ego motion of robot is not smooth.
We use Lewis-Overton line search \cite{lewis2013nonsmooth} that supports nonsmooth functions.
Readers can refer to our previous work EGO-Planner\cite{zhou2020egoplanner} for more details.

\section{Experiment Results}
\label{sec:Experiment Results}
\subsection{Implementation Details}
To validate the efficiency and accuracy of our method in trajectory optimization,
we conduct simulation comparison experiments based on convex and non-convex shape robots respectively. 
We also performed simulation experiments of our method in a 3D narrow environment based on an aerial robot.
All the simulation experiments are run on a desktop equipped with an AMD Ryzen 7 3700x 8-core CPU. 
In real world experiments, to validate our method can be applied to any-shape robots, we especially implement experiments with rectangle-shape (convex) and L-shape (non-convex) McNamee-based wheeled robots. 
All computations are performed by an onboard computer NUC with i5-1135G7 CPU. Environment information is stored by a pre-build precise point cloud map and real-time localization is obtained by a NOKOV motion capture system. 
During the optimization, all methods use the same optimization solver L-BFGS and parameters.
For safety requirements, we set the safety distance threshold to $0.1m$, which can be easily achieved in our method by expanding the RC-ESDF.
\subsection{Benchmark for Convex Shape Robot}
\label{sec:convex benchmark}
\begin{figure}[!t]
	\centering
   \vspace{0.5cm}
	\includegraphics[width=1\linewidth]{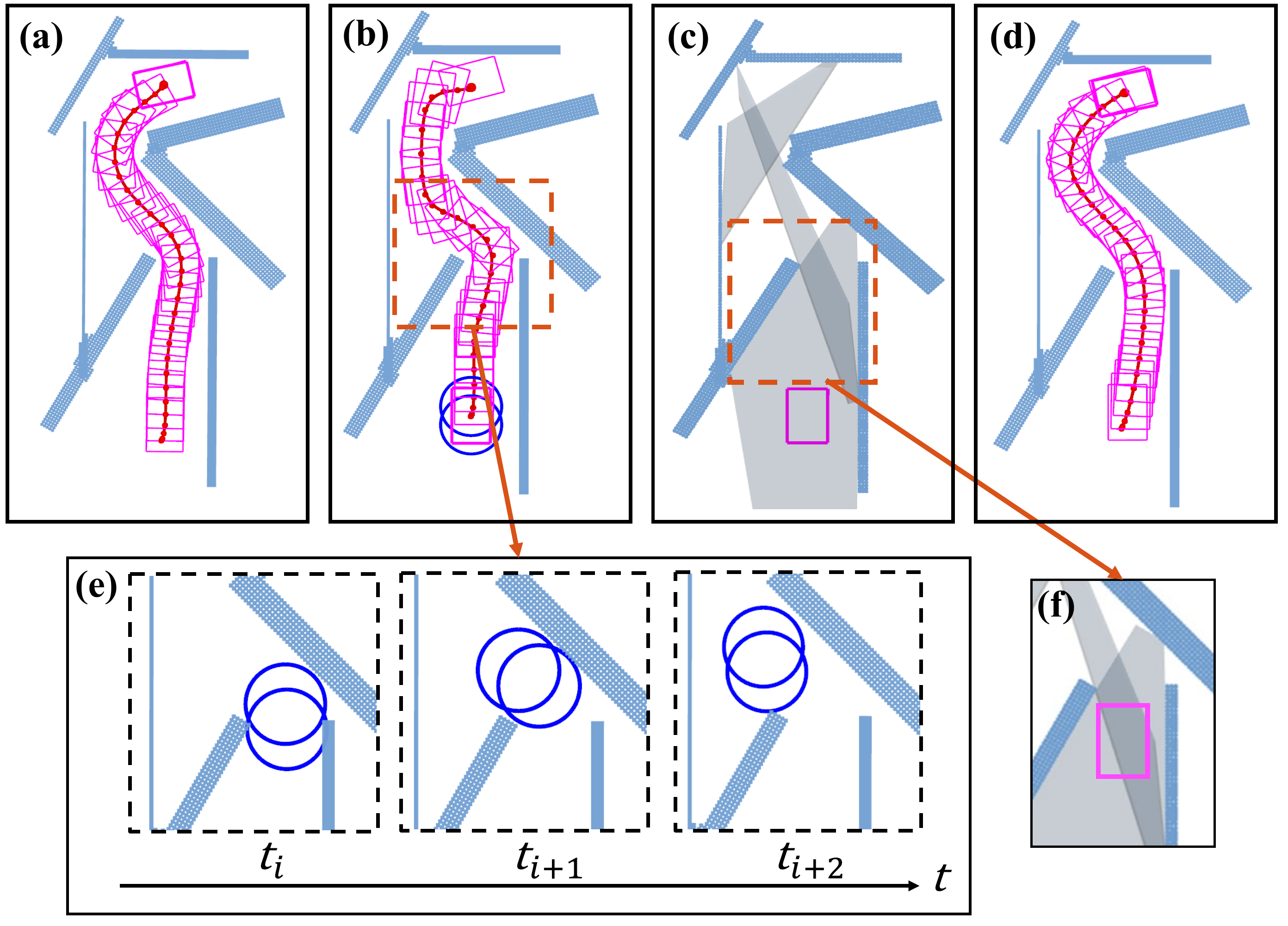}
   \captionsetup{font={small}}
	\caption{
      Comparison of the proposed method against three whole-body trajectory optimization methods to pass through narrow gaps based on convex shape robot. (a): Proposed. (b): Li's method. (c): Fast-Racing. (d): Whole-body Fast-Planner. (e): Two circles are blocked by narrow gaps of Li's method leading to abrupt yaw angle changes. (f): The intersection of two adjacent polyhedron fail to cover the whole-body of the robot, which leads to optimization failure.
	}
	\label{fig:benchmark}
\end{figure}

We benchmark our method with Env-ESDF based method Fast-Planner\cite{zhou2019robust}, Li's method \cite{li2021optimization}, and corridor-based method Fast-Racing \cite{han2021fast} based on convex shape robot.
In simulation, we set a $1.8m \times 1.2m$ rectangle-shape robot and $17m \times 10m$ narrow environment. We also provide the same reference path generated by A* as the initial value of trajectory optimization for each method. The results are shown in Fig. \ref{fig:benchmark}.

\begin{figure}[!t]
	\centering
	\includegraphics[width=1\linewidth]{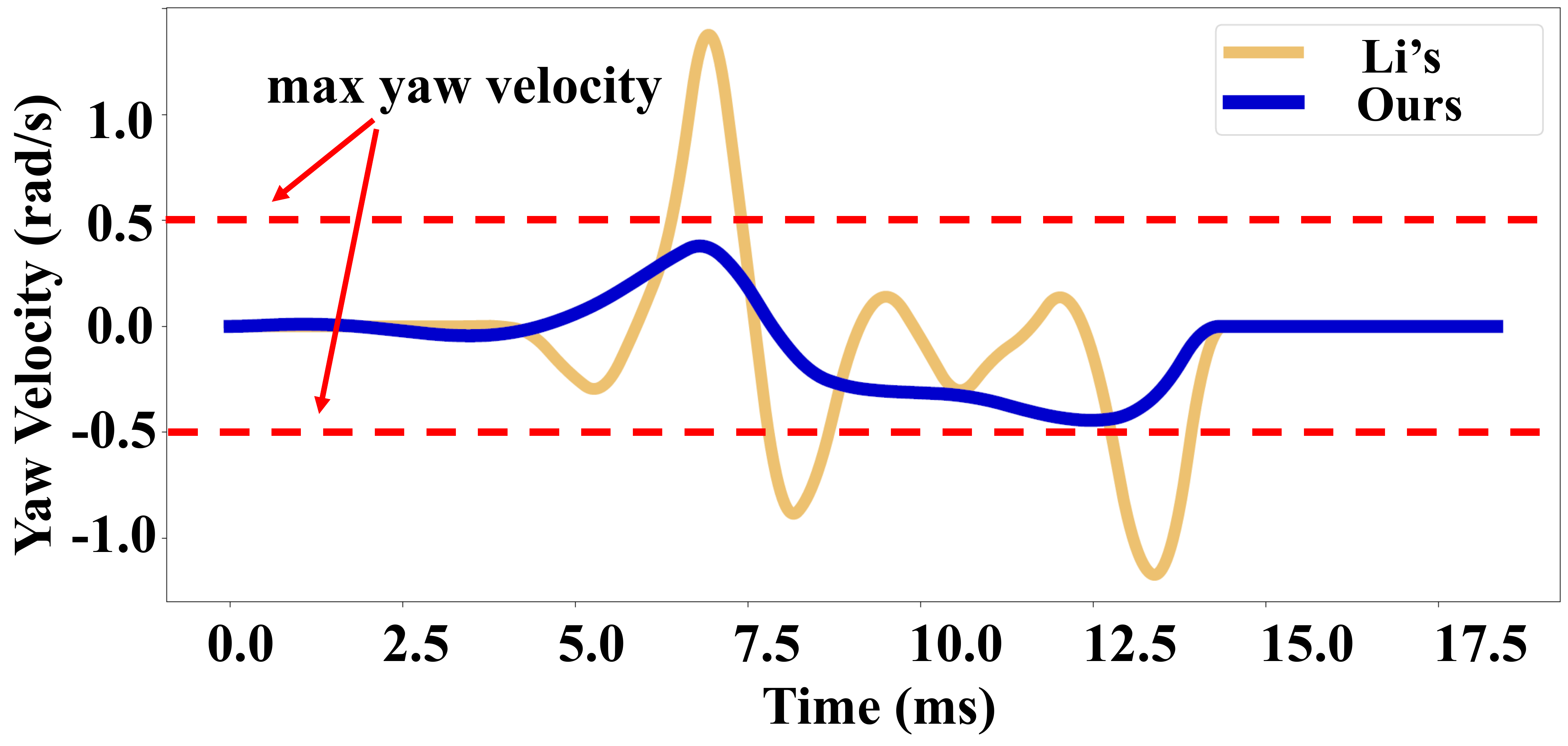}
   \captionsetup{font={small}}
	\caption{
      Comparison of Our method and Li's method for the yaw velocity of the robot.
	}
    \vspace{-1.2cm}
	\label{fig:yaw_vel_curvess}
\end{figure}

Li's method abstracts the robot's shape with two circles. However, this method is too conservative to optimize in narrow area. 
As illustrated Fig. \ref{fig:benchmark}(b) and (e), the robot is blocked by obstacles because of the low-fidelity approximation introduced by this method.
This leads to the fact that the yaw angle dramatically changes and fails to meet feasibility constraint in the optimization to enable the two circles to achieve obstacle avoidance. 
To further illustrate, we compare the yaw velocity curves of Li's and our method, as shown in Fig. \ref{fig:yaw_vel_curvess}. It clearly demonstrates that Li's method does not satisfy the dynamic feasibility constraint and has several large abrupt changes.

The result of corridor-based method Fast-Racing is shown in Fig. \ref{fig:benchmark}(c). 
The corridor is made up of grey polyhedrons. 
The result indicates that the trajectory optimization fails due to the intersection of two adjacent polyhedrons can't cover a whole-body of the robot.

\begin{figure}[!t]
	\centering
   \vspace{0.2cm}
	\includegraphics[width=0.9\linewidth]{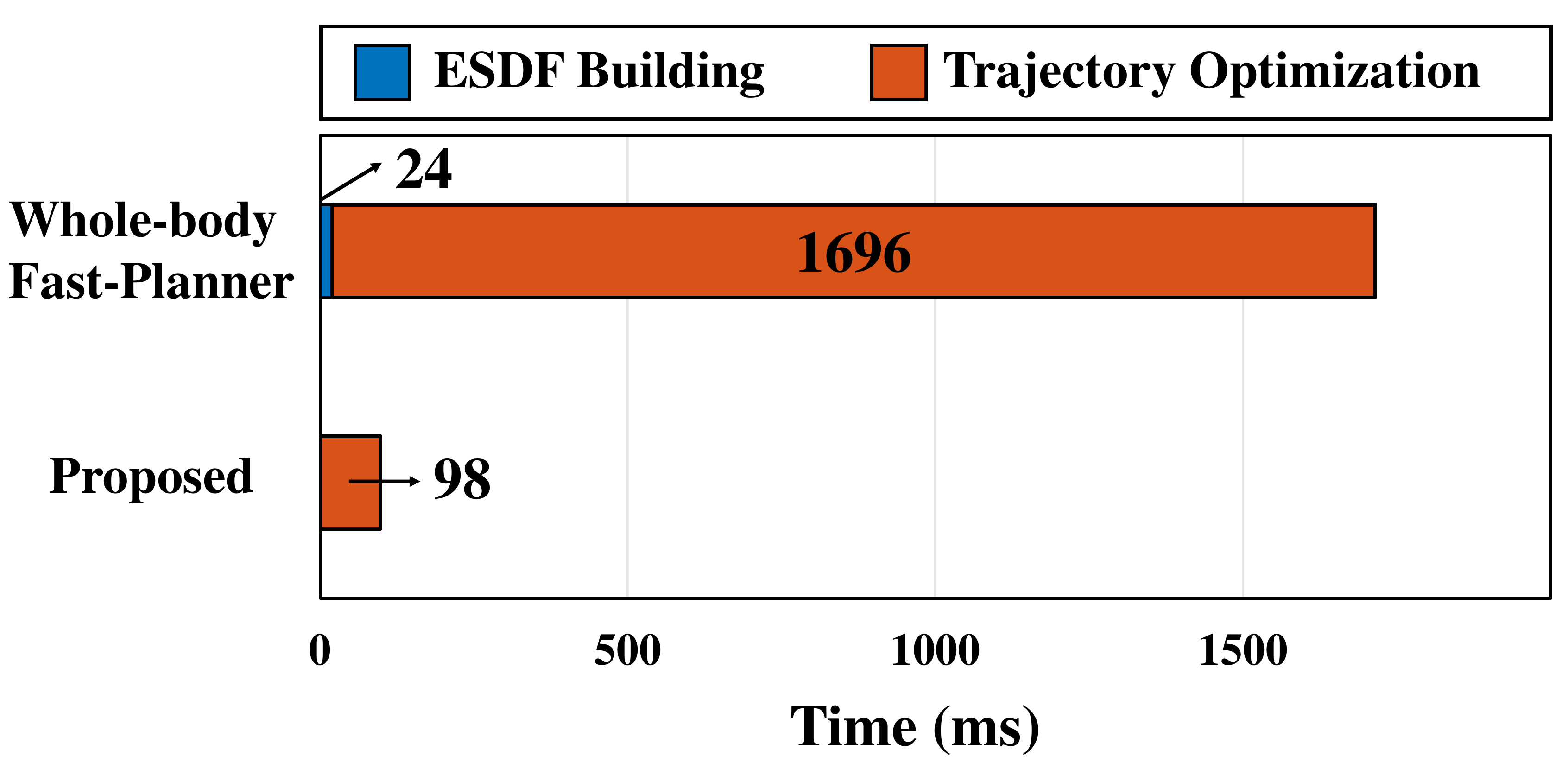}
   \captionsetup{font={small}}
	\caption{
      Comparison of proposed and whole-body Fast-Planner for ESDF building time and trajectory optimization time. Proposed method has negligible ESDF building time.
	}
    \vspace{-1.2cm}
	\label{fig:esdf_build_time}
\end{figure}

Fast-Planner treats robots as a mass point for efficient calculation in simple environments.  
An intuitive way to achieve whole-body collision evaluation is sampling points on the full shape of the robot and summing up ESDF values of these points to construct safety constraint, and we call this Env-ESDF-based method whole-body Fast-Planner (abbreviated as WBFP). To compare the differences between WBFP and proposed more fairly, we use the same trajectory optimizer L-BFGS and optimization parameters. 
As shown in Fig. \ref{fig:benchmark}(d), the trajectory of whole-body Fast-Planner has a similar effect to our method.
Then we compared the two methods in trajectory optimization time. The resolution of both the Env-ESDF in Fast-Planner and the RC-ESDF in our method is $0.1m$.
We require the two methods to repeat the trajectory optimization 20 times in the same environment, with a random start and end point and a planning length of about $15m$.
The result is illustrated in Fig. \ref{fig:esdf_build_time}, WBFP has to maintain an ESDF with a range of $16m \times 12m$ and check ESDF values of every sample points to avoid nonlinear mutations. In addition to not consuming time maintaining the ESDF, our method has less computational overhead due to lazy collision evaluation with fewer requests to ESDF for interpolation.

\subsection{Comparison of Different Size Robots}
\begin{figure}[!t]
	\centering
   \vspace{-0.3cm}
	\includegraphics[width=1\linewidth]{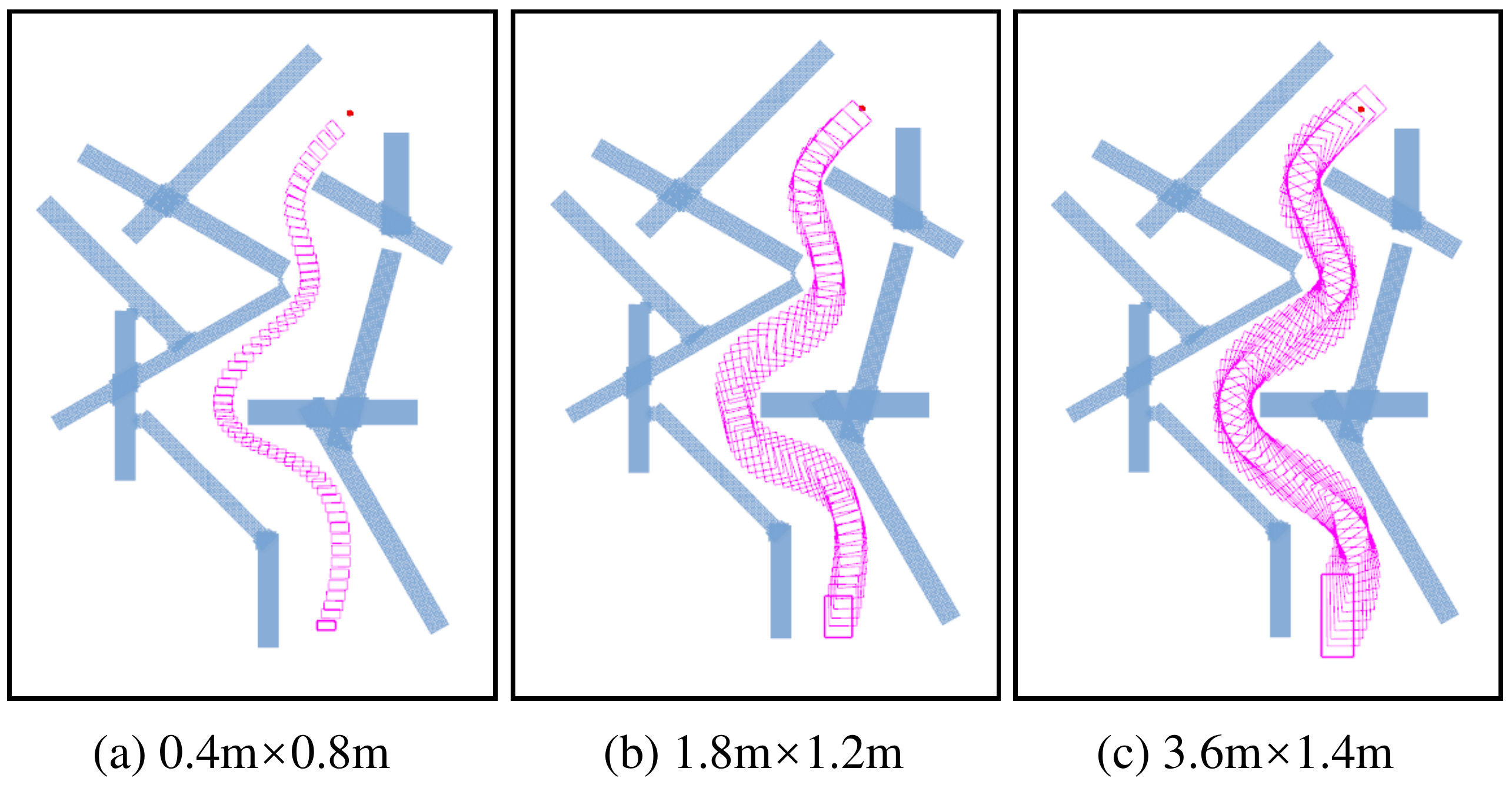}
   \captionsetup{font={small}}
	\caption{
       Trajectory optimization results for three different size robots. 
	}
    \vspace{-1.0cm}
	\label{fig:rect-size}
\end{figure}

To further demonstrate the advantages of the proposed method over whole-body Fast-Planner\cite{zhou2019robust}, we conduct experiments using rectangle-shape robots of different sizes in a $25m \times 18m$ environment. 
The initial path is still provided by A*. 
We set up robots with sizes $0.4m \times 0.8m$, $1.8m \times 1.2m$ and $3.6m \times 1.4m$ respectively, and their results are shown in Table \ref{tab:table 1} and Fig. \ref{fig:rect-size}. The total trajectory optimization and one iteration time of the proposed method are much less than that of the whole-body Fast-Planner.

Whole-body Fast-Planner requires calculating the ESDF value for each sample point at each iteration.  As the size of robot increases, the time cost of collision penalty calculation for a single iteration also increases, resulting in a longer total optimization time. Proposed method only needs to query the ESDF value of obstacle points that fall within the field at each iteration, which greatly enhances the speed of whole-body collision evaluation. 
In summary, our method achieves efficient whole-robot collision evaluation and fast trajectory optimization while accurately describing the robot shape.

\renewcommand{\arraystretch}{2.0}
\begin{table}[!t]
    \centering
    \vspace{0.2cm}
        \caption{\centering  Methods Optimization Time Comparison for Different Size Robots}
\begin{tabular}{c|c c c}
\hline
\diagbox{Methods}{Opt time}{Size} & $0.4m\times0.8m$ & $1.8m\times1.2m$ & $3.6m\times1.4m$\\
\hline
\makecell{Proposed\\(total time)} & $\textbf{0.10s}$ & $\textbf{0.31s}$ & $\textbf{1.14s}$ \\
\hline
\makecell{WBFP\\(total time)} & $1.449s$ & $13.26s$ & $11.84s$ \\
\hline
\makecell{Proposed\\(one iteration)} & $\textbf{0.11ms}$ & $\textbf{0.36ms}$ & $\textbf{1.36ms}$ \\
\hline
\makecell{WBFP\\(one iteration)} & $2.00ms$ & $10.47ms$ & $23.45ms$ \\ 
\hline
\end{tabular}
    \label{tab:table 1}
\end{table}

\subsection{Benchmark for Non-convex Shape Robot}
\begin{figure}[htbp]
	\centering
  \vspace{-0.1cm}
	\includegraphics[width=1.0\linewidth]{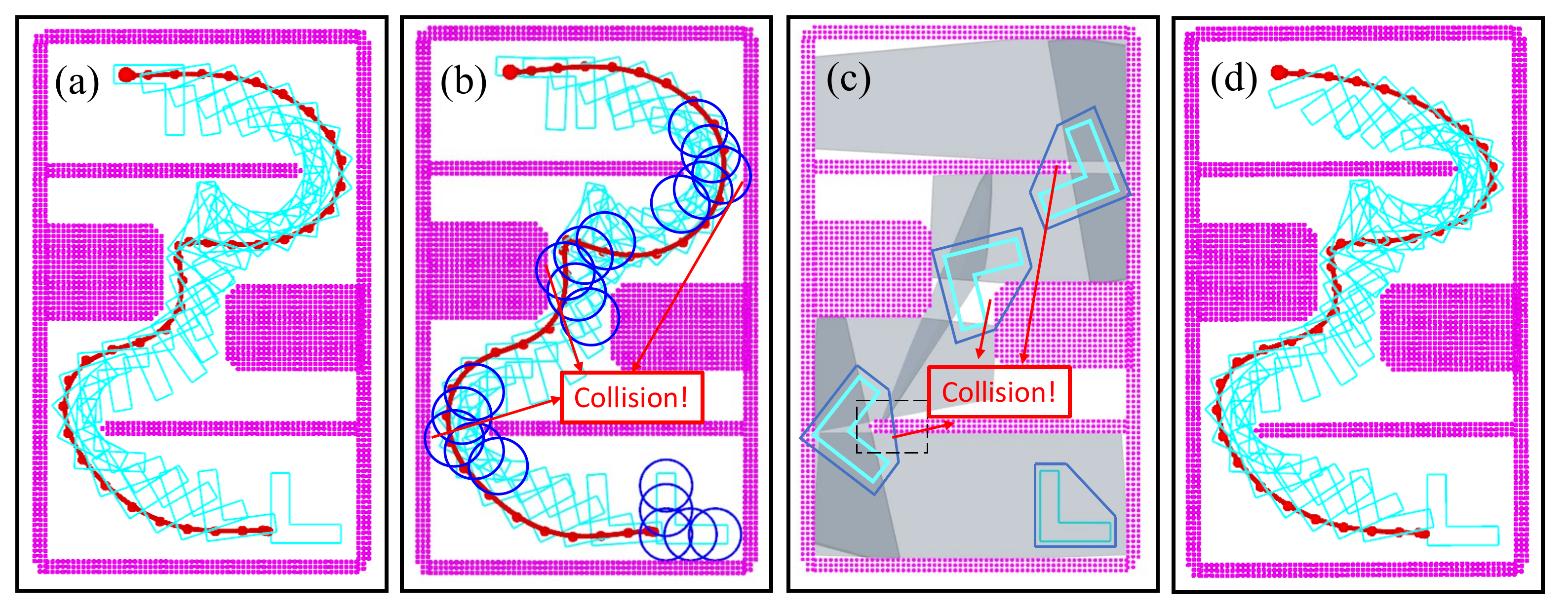}
   \captionsetup{font={small}}
	\caption{
       Comparison of the proposed method against three whole-body trajectory optimization methods to pass through narrow gaps based on non-convex shape robot. (a): Proposed. (b): Li's method. (c): Fast-Racing. (d): Whole-body Fast-Planner.
	}
    \vspace{-0.1cm}
	\label{fig:l-compare}
\end{figure}

We benchmark our method with the same methods in Sec.\ref{sec:convex benchmark} based on non-convex shape robot. We designed an L-shape robot of $1.2m$ length and $0.4m$ width and $10m \times 6m$ narrow environment. The reference path for trajectory optimization is generated by the standard implementations of RRT* from OMPL\footnote{http://ompl.kavrakilab.org/}. The comparison results are shown in Fig. \ref{fig:l-compare} and Table \ref{tab:table 2}.

Based on Li's method\cite{li2021optimization}, L-shape robot is abstracted as a combination of five circles. 
This method is conservative that the robot is blocked in a narrow area, as shown in Fig. \ref{fig:l-compare}(b), resulting in no feasible solution.
Fast-Racing\cite{han2021fast} represents the non-convex shape robot as a convex polyhedron, and strictly requires that the intersection of adjacent compositions of the corridor can contain at least one robot, as depicted in Fig. \ref{fig:l-compare}(c), leading to the safety constraint is hard to satisfy.
For Whole-body Fast-Planner\cite{zhou2019robust}, dense sampling is required at each iteration, thus increasing the total optimization time. 
In summary, our method costs a shorter trajectory optimization time while obtaining better trajectory quality.

\renewcommand{\arraystretch}{1.8}
\begin{table}[!t]
    \centering
    \vspace{0.2cm}
        \caption{\centering Methods Quantitative Comparison for Non-convex shape robot}
\begin{tabular}{c|c c c}
\hline
Methods & $opt(s)$  & $smooth(m/s^3)$ & $length(m)$  \\
\hline
Proposed  & $\textbf{0.46}$ & $0.65$ & $18.25$  \\
\hline
Li's method & $failed$ & $0.77$ & $18.55$ \\
\hline
Fast-Racing & $failed$ & $\backslash$ & $\backslash$ \\
\hline
\makecell{WBFP} & $1.29$ & $\textbf{0.60}$ & $\textbf{17.68}$ \\
\hline
\end{tabular}
\vspace{-1.3cm}
    \label{tab:table 2}
\end{table}

\subsection{Application on Aerial Robots}
\begin{figure}[htbp]
	\centering
   \vspace{-0.2cm}
	\includegraphics[width=0.7\linewidth]{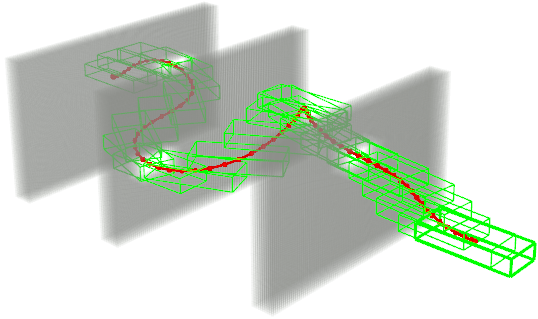}
   \captionsetup{font={small}}
	\caption{
Rectangular aerial robot traverses three narrow slits.
	}
    \vspace{-0.6cm}
	\label{fig:fly-sim1}
\end{figure}

To demonstrate the applicability of our method to three-dimensional trajectory optimization, we specifically designed a simulation experiment based on aerial robots, where the robot's trajectory is represented by $\left\{x(t), y(t), z(t), \psi(t)\right\} \in R^3 \times SO(2)$.
In simulation, a $3.6m \times 1.8m \times 0.6m$ rectangular aerial robot needs to pass through three narrow slits in succession. The result is shown in Fig. \ref{fig:fly-sim1}. The optimization time is $6.63s$, the trajectory length is $31.43m$, and the execution time is $38.12s$. In conclusion, the simulation experiment demonstrates that our method can be extended to 3D narrow environments, benefiting from accurate modeling.

\subsection{Real World Experiments}
We conduct real-world experiments in several complex environments with narrow gaps and channels. 
Thanks to the holonomic characteristics of the McNamee-based wheeled robot, the rotation of the robot is decoupled from the position, which means that the robot can pass through narrow gaps by rotating yaw angle freely. 
The reference path is generated by RRT* from OMPL. 

The rectangle-shape robot is shown in Fig. 1(a), whose length and width are $1.2m$ and $0.4m$,  respectively. 
Pink line represents the head of the robot. 
The experiment scene is depicted in Fig. 1(a) under the Bird's Eye View (BEV), and the point cloud map and trajectory after optimization is shown in Fig. 1(b). 
Narrow gaps and channels are less than $1.0m$, where the rectangle-shape robot needs to pass in a specific rotation. 
In this experiment, our method generates a $13m$ whole-body collision-free global trajectory in only $0.16s$. 
Changes of position and rotation of the robot are presented as snapshots in Fig. 1(a). 




\begin{figure}[!t]
	\centering
   \vspace{0.2cm}
	\includegraphics[width=1\linewidth]{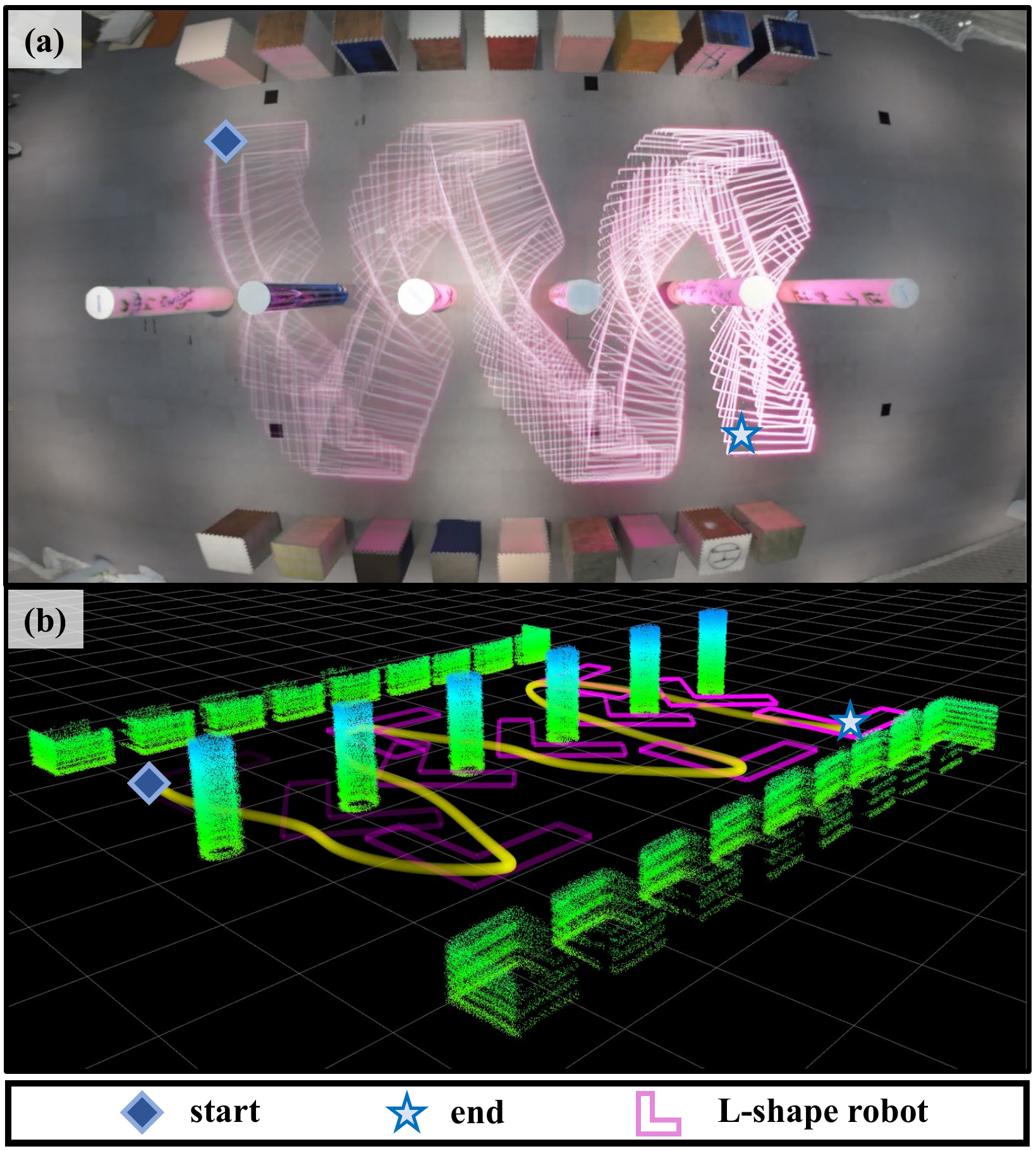}
   \captionsetup{font={small}}
	\caption{L-shape robot experiment that makes the robot pass through a line of narrow gaps by changing yaw angle like roller skating.  (a): BEV of experiment. (b): Trajectory in point cloud map.}
    \vspace{-0.7cm}
	\label{fig:roller}
\end{figure}

The L-shape robot is illustrated in Fig. 1(c). 
The longest and shortest sides of the L-shape are 1.2m and 0.4m, respectively. 
In experiment scene Fig. 1(c) and (d), the width of narrow gaps is 1.0m, which requires L-shape robot to move and rotate flexibly to pass through narrow gaps with the narrowest part of robot body. 
Another experiment scene is shown in Fig. \ref{fig:roller}, which requires the L-shape robot to cross over a line of narrow gaps like roller skating. 
The robot has to change yaw angle continuously, and keep trajectory smooth and feasible, shown in snapshot Fig. \ref{fig:roller}(a). 

The real-world experimental results prove the reliability of our method in complicated and narrow environments.
Additionally, tests on robots of different shapes verify the generality of the proposed method for both convex and non-convex shape robots.

\section{Conclusion}
In this paper, we propose a novel RC-ESDF for any-shape robot trajectory planning. Pre-building in the robot body frame, RC-ESDF does not require a real-time update.
This method ignores the information of obstacles that are out of RC-ESDF and only considers the obstacle points that fall in the field, which significantly reduces the computational overhead. 
Additionally, we jointly optimize the position and rotation of robot simultaneously to generate a collision-free whole-body trajectory for any-shape mobile vehicle robot. 
Compared with the corridor-based method and the Env-ESDF-based method, our method achieves fast trajectory optimization and explicitly considers robot's shape, which has better performance in dense environments. 
Real-world experiments validate that our method is robust to narrow environments and generic for both convex and non-convex robots.
In the future, we plan to extend our method to the SE(3) space and consider how to generate a whole-body reference path efficiently.
\bibliography{references}
\end{document}